\documentclass[conference]{IEEEtran}
\usepackage{cite}
\usepackage{graphicx}
\usepackage{pgfplots}
\usepgfplotslibrary{groupplots}
\pgfplotsset{width=\linewidth,compat=1.9}
\usepackage{url}
\usepackage{tabularx, booktabs}
\usepackage{listings}
\usepackage{xcolor}
\usepackage{makecell} 						
\usepackage[acronym]{glossaries}
\usepackage[nointegrals]{wasysym}			
\usepackage[shortlabels]{enumitem}			
\usepackage{csquotes}
\usepackage{multirow}
\usepackage{tikz}
\usetikzlibrary{arrows.meta, positioning, calc}

\definecolor{light-gray}{gray}{0.95}
\lstdefinestyle{mystyle}{
    backgroundcolor=\color{light-gray},  
    basicstyle=\ttfamily\small
}
\definecolor{codegray}{rgb}{0.5,0.5,0.5}

\newacronym{llm}{LLM}{Large Language Model}
\newacronym{strips}{STRIPS}{STanford Research Institute Problem Solver}
\newacronym{pddl}{PDDL}{Planning Domain Definition Language}
\newacronym{smt}{SMT}{Satisfiability Modulo Theories}
\newacronym{hitl}{HitL}{Human-in-the-Loop}
\newacronym{unsat}{UNSAT}{unsatisfiable}
\newacronym{sparql}{SPARQL}{SPARQL Protocol and RDF Query Language}

\begin{document}
\bstctlcite{IEEEexample:BSTcontrol} 

\title{An LLM-Based Assistance System for \\ Intuitive and Flexible Capability-Based Planning}

\author{
\IEEEauthorblockN{
    Luis Miguel Vieira da Silva,
    Nicolas König,
    Felix Gehlhoff
}
\IEEEauthorblockA{
\textit{Institute of Automation Technology, Helmut Schmidt University}\\
Hamburg, Germany\\
Email: \{miguel.vieira, felix.gehlhoff\}@hsu.hamburg\\}
}

\maketitle

\begin{abstract}
In modern industry, dynamic environments and the complexity of modular and reconfigurable resources require automated planning of process sequences. 
Capability-based planning approaches address this by automatically generating plans from semantic knowledge models that describe resource functions in a machine-interpretable form. Their practical use, however, remains limited: solver feedback, especially in the case of unsatisfiability, is difficult to interpret, and the knowledge models require adaptation as operational conditions change or requests become infeasible.
This paper presents a hybrid assistance system that augments an existing capability-based \gls{smt} planning approach with an \gls{llm}-based layer for natural-language interaction, explanation, and adaptation. Formal planning correctness remains with the symbolic planner, while the \gls{llm} layer handles natural-language access and flexible knowledge model adaptation under explicit \gls{hitl} approval. The system decomposes into four components: Capability Grounding, Symbolic Planning, Result Interpretation, and Planning Adaptation, realized as a routed agentic workflow in which a central router delegates to five specialized agents.
The system is evaluated on a modular production system across four scenario types. Of 23 test cases, 9 of 10 knowledge queries and all 4 satisfiable planning cases were handled correctly, 3 of 4 unsatisfiable cases produced concrete repair proposals, and all 5 adaptive planning scenarios resolved into satisfiable plans through iterative, user-approved knowledge model modifications. The findings confirm that combining formal planning with \gls{llm}-based assistance substantially improves accessibility and adaptability in industrial automation.
\end{abstract}


\glsresetall

\section{Introduction}
\label{sec:introduction}
Modern industrial automation systems operate in increasingly dynamic environments. Rapidly changing customer demands, shorter product life cycles, flexible batch sizes, and changing operating conditions require these systems to adapt quickly to new tasks and constraints, which is a central objective of Industry~4.0 \cite{Kagermann.08April2013}. At the same time, industrial automation environments are complex and interconnected through the modular integration of heterogeneous resources such as manufacturing machines and autonomous robots \cite{Lu.2017}. 

In such environments, process planning, i.e., the determination of a suitable sequence of functions, can no longer be regarded as a static, one-time engineering task. As manual planning becomes too labor-intensive, slow, and error-prone in such dynamic environments, automated planning approaches are required. Such approaches transform an initial system state into a desired goal state while respecting the functions and constraints of the available resources \cite{Ghallab.2004}. 

Classical symbolic planning approaches address this problem by relying on explicit formal representations of resources, functions, constraints, and goals, and using planning algorithms to compute valid plans. 
However, their practical use still requires substantial expert knowledge, particularly for formulating planning problems in dedicated formal languages and for interpreting planning failures \cite{Ghallab.2004}.
Our previous work has reduced the formulation effort by showing how capability-based knowledge models can be transformed automatically into formal planning problems \cite{Kocher.2024}. 
Yet, in realistic automation scenarios, planning problems may become unsatisfiable or require adaptation due to resource failures, modified goal specifications, or changing environmental conditions. In such cases, diagnosing the cause of failure and deriving suitable modifications to the underlying planning problem remain difficult, especially for non-expert users \cite{Ghallab.2025}.

In parallel, \glspl{llm} have demonstrated strong performance in natural-language interaction, task explanation, and complex reasoning \cite{Wei.2022}. These properties make them promising for facilitating non-expert access to formal planning systems. At the same time, purely \gls{llm}-based planning remains unsuitable for many industrial use cases because it may hallucinate and cannot be relied upon to satisfy hard logical and physical constraints consistently \cite{Wei.2025, Hao.2023}. For safety-critical and industrial applications, a hybrid architecture is therefore a promising direction: the symbolic planner remains responsible for correctness and formal consistency, while the \gls{llm} serves as an interaction, explanation, diagnosis, and adaptation layer \cite{Pallagani.2024}.

This paper follows this direction by extending our existing capability-based \gls{smt} planning approach from \cite{Kocher.2024} with an \gls{llm}-based assistance layer. The work addresses the following research questions:
\begin{enumerate}[leftmargin=*,label=\textbf{RQ\arabic*:}]
    \item How can \glspl{llm} enable natural-language interaction with capability-based planning, covering both knowledge queries and the formulation of planning problems?
    \item How can planning results for both satisfiable and unsatisfiable planning problems be explained in a way that is understandable to users?
    \item How can \glspl{llm} derive actionable adaptation suggestions from unsatisfiable solver feedback and reported environmental changes to support planning adaptation?
\end{enumerate}

The remainder of this paper is structured as follows. Section~\ref{sec:relatedWork} reviews related work on symbolic planning, \gls{llm}-based planning, and hybrid planning architectures. Section~\ref{sec:method} presents the concept and architecture of the proposed assistance system, building on our previously published capability-based planning approach. Section~\ref{sec:implementation} describes its prototypical implementation. Section~\ref{sec:evaluation} reports the evaluation results. Finally, Section~\ref{sec:conclusion} concludes the paper and outlines future work.

\section{Background and Related Work} 
\label{sec:relatedWork}
\subsection{Classical Planning Approaches}

Classical planning approaches provide the formal foundation for automated process planning by representing states, actions, and goals explicitly and by computing action sequences that achieve a desired target state \cite{Ghallab.2004}. Foundational work such as \gls{strips} established the paradigm of modeling actions through preconditions and effects, thereby enabling systematic search in state spaces \cite{Fikes.1971}. Later, standardized modeling languages such as \gls{pddl} separated domain descriptions from concrete planning instances and enabled the use of generic planning algorithms across application areas \cite{Ghallab.1998}. More recently, satisfiability-based methods have become increasingly relevant. In particular, \gls{smt}-based planning extends Boolean satisfiability by richer theories such as arithmetic and is therefore well suited for industrial planning problems with numerical constraints \cite{Cashmore.2020}.

However, defining such planning problems typically require substantial expert knowledge, while their formulation often remains manual and labor-intensive \cite{Rogalla.2017}.  

\subsection{Capability-based SMT Planning}
A relevant approach to reducing the manual effort of formal planning is our previous capability-based \gls{smt} planning method, which addresses the gap between semantic knowledge modeling and process planning by automatically generating planning problems from an existing capability model \cite{Kocher.2024}.
In this context, capabilities describe the functions of available resources together with their inputs, outputs, and constraints in a machine-interpretable form \cite{KBH+_AReferenceModelfor_15.09.2022b}. Based on this ontology-based representation, relevant planning information is extracted from the knowledge model, transformed into an \gls{smt} problem, and solved using an off-the-shelf \gls{smt} solver \cite{Kocher.2024}.
By deriving the planning problem directly from the capability model, the approach avoids the need to formulate planning instances manually in a dedicated planning language. At the same time, \gls{smt} provides a suitable formal basis for representing both logical and numerical constraints, which is particularly relevant for industrial automation scenarios.

If the generated \gls{smt} problem is satisfiable, the solver returns a valid plan in terms of selected capabilities and corresponding parameter assignments. If the problem is unsatisfiable, the solver returns a set of conflicting constraints responsible for the infeasibility. While this provides useful diagnostic information, interpreting these conflicts with respect to the underlying planning problem and capability model remains difficult, especially for non-expert users. Moreover, the approach does not provide a natural-language interface for querying capability knowledge, explaining planning outcomes, or assisting users in adapting the planning problem under changing conditions.

These limitations motivate the consideration of approaches that complement automated planning with more accessible forms of interaction, explanation, and adaptation.

\subsection{LLMs for Planning}

Recent work has explored the use of \glspl{llm} for planning due to their strong performance in natural-language understanding, reasoning, and task decomposition \cite{Wei.2025}. In contrast to classical planning approaches, \glspl{llm} can process user requests directly in natural language and generate planning-relevant outputs without requiring users to formulate problems in dedicated formal languages. This makes them attractive for improving accessibility and flexibility in planning-related tasks \cite{Pallagani.2024}.

Wang et al. propose the Describe, Explain, and Plan System, in which planning is organized as an iterative loop of four components \cite{Wang.2023}. A descriptor first translates observations from the environment into structured descriptions, an \gls{llm} then acts as an explainer that analyzes failures and evaluates states and preconditions using few-shot chain-of-thought prompting. Based on this information, the planner generates candidate subgoals that are assessed by a selector with respect to their expected success. 
However, the approach lacks formal constraint and state-transition encoding and therefore provides no guarantees of plan correctness, completeness, or precondition safety. Instead, it shifts part of the effort to costly online failure handling during execution.

Zhou et al. propose Language Agent Tree Search, which combines \gls{llm}-based reasoning and acting with Monte Carlo Tree Search to explore multiple solution paths instead of following a single sequential reasoning chain \cite{Zhou.2024}. The \gls{llm} is used both to guide search and to generate self-reflective feedback for subsequent iterations.
However, the approach still does not provide formal guarantees that the generated plans satisfy hard logical constraints. 
In addition, the considered benchmarks such as WebShop and coding tasks are not directly comparable to industrial automation scenarios with strict logical, temporal, and resource-related requirements.

Silver et al. use GPT-4 as a generalized planner that synthesizes executable Python programs from \gls{pddl} domains and few-shot examples \cite{Silver.2024}. The generated programs are validated externally with a \gls{pddl} validation tool and iteratively repaired if syntax or semantic errors occur, which improves inspectability and verifiability. 
However, the evaluation is limited to rather simple generalized-planning domains for which generalized solutions could largely be provided manually. Moreover, the method depends strongly on meaningful \gls{pddl} identifiers, suggesting that performance relies not only on formal structure but also on natural-language cues and possible pretraining bias. 

Overall, these works show that \glspl{llm} can usefully support planning-related reasoning, task interpretation, and user interaction. At the same time, when used as the main planning component, they remain limited by hallucinations, missing formal guarantees, and unreliable handling of hard constraints \cite{Hao.2023, Wei.2025}. This is particularly problematic in industrial automation, where planning results must satisfy explicit logical and numerical constraints and remain consistent with the underlying system model. Accordingly, recent work increasingly motivates hybrid approaches in which \glspl{llm} complement rather than replace formal planning back ends, for example, by supporting interaction, explanation, or adaptation while correctness is ensured by symbolic methods \cite{Pallagani.2024}.

\subsection{Hybrid Planning Approaches}

To address the limitations of purely \gls{llm}-based planning, recent work increasingly combines \glspl{llm} with classical planning methods. In these hybrid approaches, \glspl{llm} are typically used for natural-language understanding, translation, explanation, or intermediate reasoning, while formal planners or solvers remain responsible for computing correct plans. 

Liu et al. propose \mbox{LLM+P}, a framework in which an \gls{llm} translates natural-language task descriptions into formal \gls{pddl} problem specifications that are then solved by a classical planner \cite{Liu.2023}. This combines the accessibility of natural-language interaction with the correctness of symbolic planning. However, the approach assumes a fixed planning domain and focuses on generating problem instances rather than adapting the domain description itself. As a result, it is less suitable for dynamic automation scenarios in which not only goals but also the underlying planning basis may need to be modified under changing conditions.

Ye et al. present SATLM, which uses an \gls{llm} to parse natural-language reasoning problems into formal constraints and delegates the actual solving to a symbolic satisfiability solver \cite{Ye.2023}. Pan et al. pursue a similar direction with Logic-LM, which structures reasoning into problem formulation, symbolic reasoning, and result interpretation \cite{Pan.2023}. Natural-language problems are translated into symbolic representations, solved by a deterministic symbolic reasoner, and translated back into natural language. A self-refinement mechanism further uses solver feedback to iteratively improve the generated formulation. Both approaches demonstrate the benefits of combining \glspl{llm} with formal reasoning back ends. However, they are evaluated on rather small and simple tasks that are not directly comparable to industrial automation scenarios and do not focus on adapting the planning basis under changing conditions or on handling unsatisfiable planning problems.

Jha et al. combine \glspl{llm} with an \gls{smt} solver by letting the \gls{llm} generate an initial solution or formalization, which is then verified and iteratively corrected using solver-generated counterexamples \cite{Jha.2023}. However, the \gls{llm} remains the core component for solution generation, while the solver mainly serves verification and repair. Moreover, the approach does not support users in understanding results or deriving concrete adaptations of the planning basis.

Hirsch et al. analyze the planning weaknesses of \glspl{llm}, particularly their difficulties in tracking world states and identifying applicable actions, and use these findings to develop Similarity-based Planning (SimPlan) \cite{Hirsch.2024}. In this hybrid approach, classical planning components provide state and action information, while the \gls{llm} is used as a heuristic for action selection. However, the approach assumes that the relevant planning knowledge is already available in a formal symbolic representation, leaving the challenge of semantic grounding unresolved. Moreover, although the work analyzes model failures in detail, it does not provide a user-oriented explanation or assistance layer for making such failures actionable.

Overall, hybrid approaches demonstrate that \glspl{llm} are most effective in planning settings when they complement rather than replace formal planning methods. Existing work mainly focuses on problem translation, heuristic support, or consistency repair, while support for natural-language access to planning knowledge, understandable explanation of unsatisfiable results, and concrete guidance for adapting the planning problem remains limited. This gap is particularly relevant for industrial planning, where non-expert users must interact with formal planning systems under changing requirements and disturbances.

\section{Method -- Hybrid Planning Approach}
\label{sec:method}
This work proposes an \gls{llm}-based assistance system for capability-based \gls{smt} planning in industrial automation. The method augments, rather than replaces, the underlying formal planner in order to improve accessibility for non-expert users. In particular, it addresses three objectives:
\begin{enumerate}[leftmargin=*, label=(\roman*)]
    \item  natural-language access to capability knowledge and planning functionality,
    \item understandable explanation of planning outcomes including unsatisfiable results, and
    \item user-guided support for adapting the planning basis under changing conditions.
\end{enumerate}
To achieve this, the method combines a formal planning back end for correctness and constraint satisfaction with an \gls{llm}-based assistance layer for interaction, explanation, and adaptation support.

\subsection{System Overview}
\label{sec:overview}

\begin{figure*}[t]
\centering
 \includegraphics[width=\linewidth]{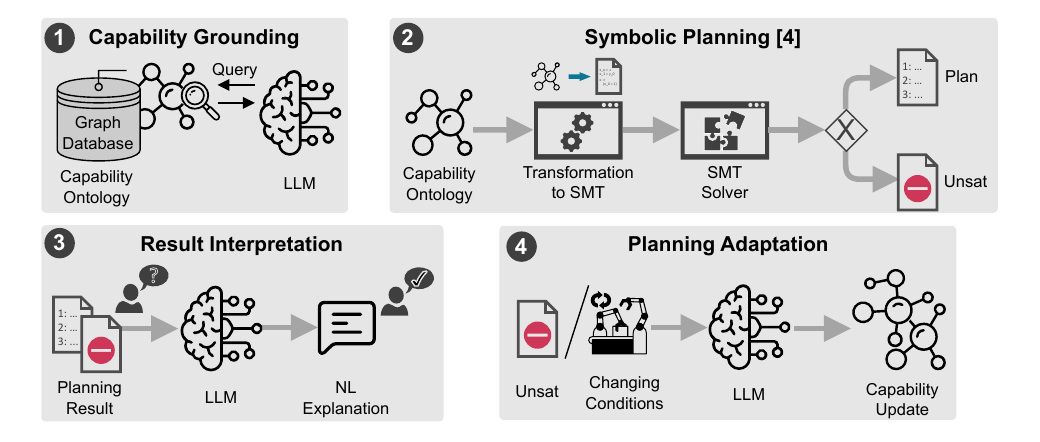}
\caption{Overview of the proposed hybrid planning assistance system, comprising four components: \emph{Capability Grounding} interprets user requests and grounds them in the capability knowledge; \emph{Symbolic Planning} transforms the capability model into an \gls{smt} problem and solves it; \emph{Result Interpretation} translates the outcome into natural language; and the \emph{Planning Adaptation} handles unsatisfiable results and changed conditions through knowledge graph updates.}
\label{fig:architecture}
\end{figure*}

The system architecture comprises four components, drawing on the hybrid decomposition proposed in \cite{Pan.2023}: \emph{Capability Grounding}, \emph{Symbolic Planning}, \emph{Result Interpretation}, and the \emph{Planning Adaptation}. Fig.~\ref{fig:architecture} gives a schematic overview.

\emph{Capability Grounding} is the entry point for all user interaction. The user's natural-language request is interpreted and matched against the available capability knowledge. The provided capabilities of available resources are already modeled in the capability ontology and form the domain description. Depending on the intent, the system either answers a knowledge query directly from the graph or identifies candidate required capabilities -- describing a specific goal for the industrial automation system -- corresponding to the user's goal, which the user then confirms, thereby establishing the planning problem. \emph{Symbolic Planning} takes the confirmed required capability together with the provided capabilities and automatically transforms them into an \gls{smt} problem and solves it by using a solver as presented in our approach from \cite{Kocher.2024}. If the problem is satisfiable, a plan is returned in terms of selected capabilities and parameter assignments. Otherwise, the solver returns a minimal unsatisfiable set of constraints. \emph{Result Interpretation} translates the outcome into a natural-language explanation for the user, covering both successful plans and infeasibility diagnoses.
The \emph{Planning Adaptation} is entered in two situations. First, when planning yields an unsatisfiable result, the unsatisfiable set of constraints is analyzed to determine the root cause of infeasibility, a concrete adaptation to the modeled capabilities is derived and presented to the user for approval, and planning is re-attempted with the updated knowledge graph. This loop continues until a satisfiable plan is found or the conflict is classified as unresolvable. Second, when a changed situation is reported, the affected provided capabilities must first be updated in the knowledge graph before planning can be re-attempted. In both cases, the knowledge graph is updated with explicit user validation, ensuring that the capability model reflects the current state of the system before a new plan is sought. This process is distinct from the self-refinement mechanism in \cite{Pan.2023}, which corrects errors in problem formulation; here, the problem formulation is correct but the capability model itself must be adapted to match changed or infeasible conditions.

\subsection{Agent Workflow}
\label{sec:agents}

The decomposition into four components described above naturally imposes different requirements on the underlying \gls{llm} usage. For example, Capability Grounding requires querying structured graph data and interpreting diverse user intents, while symbolic Planning requires invoking an external solver and interpreting its output.  
The subtasks of the four components differ in their required tools, reasoning styles, and context -- which makes a decomposition into specialized \gls{llm} agents a natural fit: each agent can be given a focused system prompt, a targeted tool set, and only the state information it actually needs, rather than being confronted with the full complexity of the system in every call \cite{Li.2024}.

The key design decision within this agent-based realization concerns the coordination strategy. Three patterns were considered. A \emph{monolithic} architecture uses a single \gls{llm} with one large prompt that must handle all possible user intents, tool calls, and reasoning tasks. In practice this leads to prompt overload: heterogeneous instructions interfere with each other, the relevant context for any given subtask is diluted by irrelevant information from others, reliability degrades as prompt complexity grows, and traceability suffers -- it becomes difficult to understand which part of the prompt drove a given output or where a failure originated. A \emph{fully autonomous} multi-agent system gives each agent the freedom to decide dynamically which sub-agents to invoke and in what order. While this maximizes flexibility, the execution path becomes difficult to forecast: it is unclear in advance which steps will be performed, in what sequence, and whether required \gls{hitl} approval points will be reached before the capability model is modified. A \emph{routed} workflow addresses both shortcomings. The overall control flow follows a defined structure that can be anticipated and traced, while individual agents retain full flexibility in how they use their tools to accomplish their specific subtask. A central Router Agent classifies the user intent and delegates to the appropriate specialist, while \gls{hitl} checkpoints are placed at fixed, well-known positions in the workflow, ensuring that no capability model modification is applied without explicit user approval.

\begin{figure*}[h!]
\centering
\includegraphics[width=\linewidth]{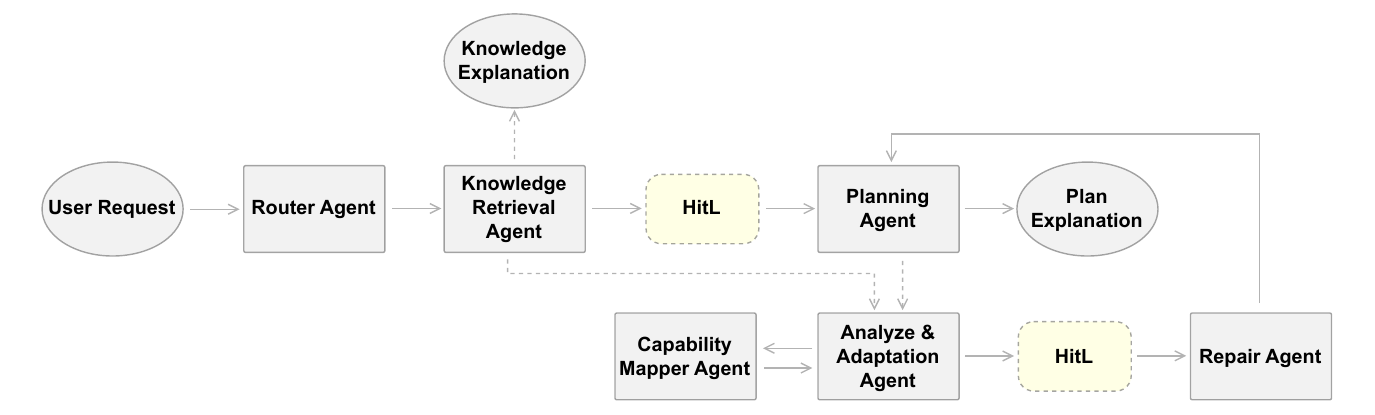}
\caption{Agentic workflow illustrating the three interaction paths: a knowledge query is answered directly by the Knowledge Retrieval Agent; a planning request proceeds through capability retrieval and the Planning Agent, with the unsatisfiable path triggering the Analyze and Adaptation Agent and the Repair Agent in a replanning loop; and a runtime failure report triggers capability identification, an adaptation proposal, and a knowledge graph update before replanning. \textit{HitL} markers indicate human validation checkpoints.}
\label{fig:workflow}
\end{figure*}

To define the agents required for this routed workflow, the overall process flow of the system was analyzed and decomposed into distinct process types. Four process types were identified: (i) \emph{intent classification}, which must determine what kind of request the user has issued before any other processing can begin, (ii) \emph{knowledge retrieval}, which queries the capability knowledge graph to obtain structured information about available resources and their capabilities, (iii) \emph{planning and explanation}, which invokes the formal solver and translates its output into language understandable to the user, and (iv) \emph{constraint analysis and adaptation}, which reasons over infeasible results and derives concrete proposals for modifying the capability model. Each of these process types is sufficiently distinct in its required tools, inputs, and reasoning style to warrant dedicated agents. The workflow therefore comprises six specialized agents:

\begin{itemize}[leftmargin=*]
    \item \emph{Router Agent}: classifies user input into one of three intent categories -- knowledge query, planning request, or runtime failure report -- and routes control to the appropriate workflow branch.
    \item \emph{Knowledge Retrieval Agent}: retrieves structured information from the capability knowledge graph, including descriptions and parameter constraints of available provided capabilities.
    \item \emph{Capability Mapper Agent}: establishes the connection between formal constraint or failure information and the capability descriptions stored in the knowledge graph. 
    This mapping is essential not for the planning step itself -- which operates directly on the formal model -- but is required in two subsequent situations: First, when the solver returns an unsatisfiable result, the Analyzer needs to know which capability a given constraint refers to in order to reason about what needs to change in the model by comparing the descriptions and parameter structures of required and provided capabilities. Second, when a changed condition is reported, the system must identify which capabilities in the knowledge graph are affected before it can be updated.
    \item \emph{Planning Agent}: triggers the \gls{smt} planner, receives the result, and translates the plan or infeasibility outcome into natural language for the user.
    \item \emph{Analyze and Adaptation Agent}: interprets the unsatisfiable set of constraints or reported runtime changes using chain-of-thought reasoning and derives a concrete adaptation proposal for the capability model.
    \item \emph{Repair Agent} executes the approved modifications to the capability knowledge graph. Once the user has confirmed a proposed adaptation via a \gls{hitl} checkpoint, the \emph{Repair Agent} applies the necessary changes and prepares the updated model for replanning.
\end{itemize}

A shared state object persists intermediate results across agents within a session. \gls{hitl} control points are placed at three stages: (1) confirmation of the target required capability before planning is initiated, (2) approval of proposed capability model adaptations in response to an unsatisfiable result, and (3) confirmation of capability model changes when a runtime failure is reported. These checkpoints ensure that no modification to the capability model is applied without explicit user approval. 

The workflow supports three interaction paths, illustrated in Fig.~\ref{fig:workflow}. For a \emph{knowledge query}, the Router Agent routes the request directly to the Knowledge Retrieval Agent, which answers from the capability knowledge graph and returns a natural-language response to the user. For a \emph{planning request}, the Router Agent routes to the Knowledge Retrieval Agent to retrieve suitable required capabilities, which the user confirms via the first \gls{hitl} checkpoint, before the Planner Agent invokes the solver. If the result is satisfiable, the plan is returned to the user in natural language. If it is unsatisfiable, the Analyze and Adaptation Agent analyzes the constraint set, the Capability Mapper Agent then establishes the connection to  provided capabilities and required capability, derives an adaptation proposal, and the user approves it via the second \gls{hitl} checkpoint before the Repair Agent updates the knowledge graph and replanning is attempted. For a \emph{runtime failure report}, the Router Agent routes to the Analyze \& Adaptation Agent, which routes to Capability Mapper Agent to identify the affected capabilities; the Analyze and Adaptation Agent proposes the necessary capability model update, the user confirms via \gls{hitl} checkpoint, the Repair Agent applies the change, and replanning is initiated.

\subsection{Planning Adaptation Cycle}
\label{sec:adaptation}

When the planner returns an unsatisfiable result or a changed operational condition is reported, the system enters the \emph{Planning Adaptation Cycle}. Rather than terminating or returning a bare failure message, the system attempts to identify why the current capability model cannot satisfy the request and to propose a concrete, user-approved modification.

The cycle proceeds in four steps. First, the Analyze and Adaptation Agent examines the unsatisfiable constraint set -- or the reported change -- using chain-of-thought reasoning to determine the root cause of infeasibility. Second, it derives a concrete adaptation proposal: a targeted modification to the capability knowledge graph that would resolve the identified conflict. Third, the proposal is presented to the user for approval (\gls{hitl}), ensuring that no change is applied autonomously. Fourth, once approved, the Repair Agents updates the knowledge graph and planning is re-attempted. This loop continues until a satisfiable plan is found or the conflict is classified as unresolvable given the available resources.

The cycle is entered from two distinct situations. In the \gls{unsat} case, the constraint analysis drives the adaptation proposal directly from the solver output. In the changed-condition case, the user reports a runtime event (e.g., a resource becoming unavailable), the relevant provided capabilities are removed or updated in the knowledge graph with user confirmation, and planning is then re-attempted with the updated model. In both cases, the knowledge graph is updated before re-planning, ensuring that the capability model reflects the current state of the system.

\section{Implementation}
\label{sec:implementation}
The system is implemented in Python using LangGraph\footnote{https://www.langchain.com/langgraph}. as the workflow orchestration framework. LangGraph represents the agentic workflow as a directed graph in which nodes correspond to agents and conditional edges encode the routing logic determined by the shared state. Each agent is realized as a ReAct agent \cite{Yao.2023}: it receives the current shared state and iteratively reasons about which of its available tools to call, executes them, and incorporates the results before producing a final output that is written back to the shared state. The overall control flow, however, follows the fixed graph structure -- routing decisions are determined by the graph edges, not by the agent itself, keeping the execution path predictable and easily inspectable.

\begin{figure}[t]
\centering
\includegraphics[width=\linewidth]{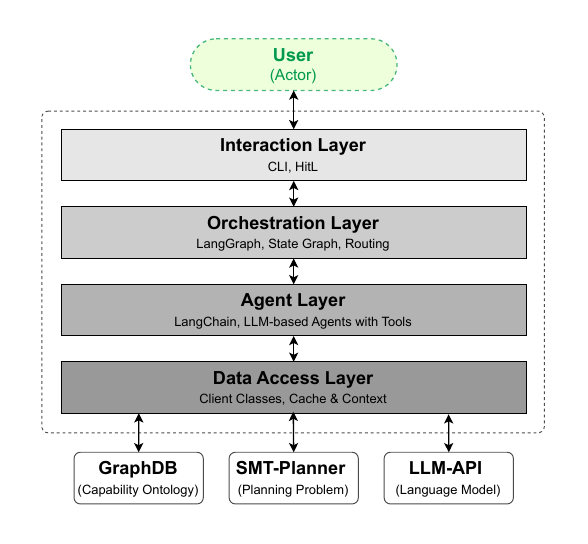}
\caption{Four-layer implementation architecture of the assistance system, comprising an interaction layer (CLI, \gls{hitl} logic), an orchestration layer (LangGraph state graph and routing), an agent layer (LangChain-based \gls{llm} agents with tools), and a data-access layer (client classes, caching) connecting to GraphDB (capability ontology), the \gls{smt} planner, and the \gls{llm} API.}
\label{fig:implementation}
\end{figure}

Fig.~\ref{fig:implementation} shows the four-layer implementation architecture: an \emph{interaction layer} handling user I/O and \gls{hitl} logic via a CLI; an \emph{orchestration layer} realizing the LangGraph state graph, shared state, and routing; an \emph{agent layer} executing the \gls{llm}-based ReAct agents via LangChain; and a \emph{data-access layer} providing client interfaces to the three external systems.

Tool definitions are scoped per agent. The \emph{Knowledge Retrieval Agent} is equipped with \gls{sparql} SELECT and ASK query tools to retrieve structured information from the capability knowledge graph hosted in a GraphDB triple store. The \emph{Capability Mapper Agent} compares capability descriptions and parameter structures of required and provided capabilities to reduce the search space for the Analyze and Adaptation Agent. The \emph{Planner Agent} invokes our planning approach from \cite{Kocher.2024} as an external tool and receives either a valid plan or an unsatisfiable constraint set. The \emph{Router Agent} performs few-shot prompt classification with no external tool calls. The \emph{Analyze and Adaptation Agent} employs chain-of-thought prompting as introduced in \cite{Wei.2022} to reason over infeasible constraints or reported runtime changes and derive a concrete adaptation proposal. The \emph{Repair Agent} is equipped with \gls{sparql} INSERT and DELETE update tools to apply approved modifications to the knowledge graph. All agents use GPT-4o mini as the underlying \gls{llm}. The Analyze and Adaptation Agent uses GPT-4o to meet the higher reasoning demands of multi-step constraint analysis and adaptation derivation. The temperature parameter is set to zero to increase reproducibility and reliability. All \gls{llm} calls enforce structured outputs via a JSON schema defined in each agent's system prompt, ensuring that intermediate results are machine-readable and can be reliably passed between agents through the shared state.

\section{Evaluation}
\label{sec:evaluation}
\subsection{Experimental Setup}
\label{sec:setup}

The system is evaluated using a capability model of a modular production system available in our laboratory. The model encodes the provided capabilities of resources as OWL individuals conforming to the CSS capability reference model \cite{KBH+_AReferenceModelfor_15.09.2022b}, including input, output, and constraint parameters. The modeled capabilities and test scenarios are publicly available\footnote{https://github.com/hsu-aut/MPS500-Capabilities}.

Evaluation scenarios are organized into four categories that together cover all three research questions:

\begin{itemize}[leftmargin=*]
    \item \emph{Knowledge Query (KQ)}: natural-language questions about available capabilities and resources, answered by querying the knowledge graph without triggering the planner (RQ1).
    \item \emph{SAT Planning}: planning requests for which a valid plan exists; the system must identify the correct required capability, invoke the solver, and produce a correct natural-language explanation of the resulting plan (RQ1, RQ2).
    \item \emph{UNSAT Planning}: planning requests that are infeasible under the current capability model; the system must recognize infeasibility and provide an understandable explanation of the cause (RQ2).
    \item \emph{Adaptive Planning (AP)}: planning requests that begin as infeasible but can be resolved through iterative knowledge graph adaptation; the system must propose, obtain user approval for a proposed adaptation, and apply modifications until a satisfiable plan is found (RQ3).
\end{itemize}

A total of 23 test cases were evaluated: ten knowledge queries, four SAT scenarios, four UNSAT scenarios, and five adaptive planning scenarios. Each test case specifies a natural-language user request and a reference outcome. SAT, UNSAT, and AP scenarios were each repeated five times to assess reproducibility. Outcomes are assessed qualitatively against scenario-specific criteria: KQ cases on completeness, ontological correctness against the underlying graph, and understandability; SAT cases on step completeness, parameter transparency, and understandability of the natural-language plan; UNSAT cases on identification of the affected capabilities, description of the conflicts, and concreteness of the repair proposal. KQ, SAT, and UNSAT use a three-level scale (fully, partially, not satisfied). AP uses a binary scale assessing whether the SPARQL repair was applied and whether replanning succeeded. A case is counted as successful when all its criteria are fully satisfied across all repetitions.

\subsection{Results}

Table~\ref{tab:results} summarizes the results across scenario categories.
\begin{table}[t]
\caption{Evaluation results per scenario category. \emph{Successful} counts cases in which all scenario-specific criteria (defined in Section~\ref{sec:setup}) were fully satisfied across all repetitions.}
\label{tab:results}
\centering
\small
\begin{tabular}{lccc}
\toprule
\textbf{Scenario} & \textbf{Cases} & \textbf{Repetitions} & \textbf{Successful} \\
\midrule
Knowledge Query   & 10 & 1 & 9 \\
SAT Planning      &  4 & 5 & 4 \\
UNSAT Planning    &  4 & 5 & 3 \\
Adaptive Planning &  5 & 5 & 5 \\
\bottomrule
\end{tabular}
\end{table}

Knowledge queries were fully satisfied in 9 of 10 cases. One result was only partial correct because it omitted the explicit parameter range justifications, leaving the answer technically correct but incomplete; no hallucinations were observed in any of the 10 cases, confirming reliable grounding in the capability ontology (RQ1).

All 4 SAT planning cases were fully satisfied across all 5 repetitions. The Planner Agent consistently translated structured solver output into clear natural-language explanations, including correctly sequencing multi-step plans such as transport followed by drilling, and listing the relevant parameter assignments alongside each step (RQ1, RQ2).

UNSAT diagnoses were fully satisfactory in 3 of 4 cases (RQ2). As a representative example, a user request triggered the system to identify two independent conflicts in a single response: \emph{Conflict~1} -- ``the required output product depth of 2\,mm is outside the acceptable range of 5\,mm to 10\,mm for the drilling module'', with the proposed repair of adjusting the depth to 5\,mm; and \emph{Conflict~2} -- ``the required station ID is set to 15, which does not match the provided station ID of 3'', with the proposed repair of aligning the required station ID to 3. This same case was also the only one with reproducibility variance: in 2 of 5 repetitions the proposal aligned the transport \emph{target} station to the start station (3) instead of the conveyor's reachable end station (7) -- a repair that is formally consistent but causes the planner to drop the transport step and return a plan that misses the user's transport intent. The remaining UNSAT cases yielded near-verbatim identical responses across all five repetitions.

All 5 adaptive planning cases were successful: approved modifications were applied via SPARQL and replanning yielded satisfiable results in every run. Case~4 -- corresponding to the multi-conflict UNSAT scenario above -- required two sequential repair iterations, with the second UNSAT-core surfacing only after the first repair had been applied. Case~5 exercised the runtime-failure path: in response to the report \emph{``The conveyor is defective.''}, the Capability Mapper Agent identified the matching provided capability and the Repair Agent removed it from the knowledge graph, after which a feasible plan using alternative resources was generated (RQ3).

Overall, the results indicate that the hybrid architecture successfully combines formal planning correctness with accessible natural-language interaction. The \gls{hitl} approval mechanism ensured that no unauthorized modification to the capability model was applied in any test case.

\section{Conclusions and Future Work}
\label{sec:conclusion}
This paper presented a hybrid \gls{llm}-based assistance system for capability-based \gls{smt} planning in industrial automation. The system augments an existing formal planning approach \cite{Kocher.2024} with a natural-language interaction and explanation, and an adaptation layer realized as a routed agentic workflow. 

The combination of the Router Agent, which classifies user intent, and the Knowledge Retrieval Agent, which grounds the user's request in the structured capability ontology, shows that users can query capability knowledge and initiate planning through natural language (RQ1). 
Both satisfiable plans and infeasible results are translated into understandable explanations, with chain-of-thought reasoning over the \gls{unsat} core enabling structured diagnosis of constraint conflicts (RQ2). Planning adaptation under changing or infeasible conditions is achieved as an iterative loop in which the Analyze and Adaptation Agent proposes targeted knowledge graph modifications, the user approves them via \gls{hitl} checkpoints, and planning is re-attempted with the updated model (RQ3).

Overall, the results confirm that a hybrid architecture -- combining the formal correctness of \gls{smt}-based planning with the natural-language capabilities of \glspl{llm} -- can substantially improve the accessibility and adaptability of capability-based planning systems in industrial automation settings.

Several directions remain for future investigation. First, the current knowledge retrieval relies on handcrafted \gls{sparql} tools. Replacing these with an MCP server, exposing structured graph access, would increase flexibility in querying the capability knowledge graph and decouple the agent layer from the specific triple store implementation, even if at the cost of potentially higher token usage. 
Second, runtime failure reports are currently triggered by user-initiated natural-language observations. Tighter integration with live sensor data streams and automated anomaly detection would enable proactive identification of capability model inconsistencies, with the remaining challenge being the reliable translation of detected anomalies into targeted knowledge graph modifications. 
Finally, the current evaluation is limited to a single use case. A larger-scale evaluation involving real non-expert users would provide stronger evidence for usability and generalizability.

\section*{Acknowledgment}
This research is funded by dtec.bw – Digitalization and Technology Research Center of the Bundeswehr as part of the project RIVA. dtec.bw is funded by the European Union – NextGenerationEU

\bibliographystyle{./bibliography/IEEEtran}
\bibliography{./bibliography/references} 

@IEEEtranBSTCTL{IEEEexample:BSTcontrol,
CTLuse_forced_etal       = "yes",
CTLmax_names_forced_etal = "3",
CTLnames_show_etal       = "2" }

@misc{Kagermann.08April2013,
 author = {Kagermann, Henning and Wahlster, Wolfgang and Helbig, Johannes},
 date = {2013},
 title = {{Recommendations for implementing the strategic initiative INDUSTRIE 4.0: Final report of the Industrie 4.0 Working Group}},
 editor = {acatech}
}

@article{Lu.2017,
 author = {Lu, Yang},
 year = {2017},
 title = {{Industry 4.0: A survey on technologies, applications and open research issues}},
 pages = {1--10},
 volume = {6},
 issn = {2452-414X},
 journal = {{Journal of Industrial Information Integration}},
 doi = {10.1016/j.jii.2017.04.005}
}

@book{Ghallab.2004,
 author = {Ghallab, Malik and Nau, Dana S. and Traverso, Paolo},
 year = {2004},
 title = {{Automated planning: Theory and practice}},
 address = {Amsterdam},
 publisher = {{Elsevier/Morgan Kaufmann}},
 isbn = {1281007218},
 series = {{The Morgan Kaufmann Series in Artificial Intelligence}}
}

@incollection{Kocher.2024,
 author = {K{\"o}cher, Aljosha and Vieira da Silva, Luis Miguel and Fay, Alexander},
 title = {{Automated Process Planning Based on a Semantic Capability Model and SMT}},
 booktitle = {{AAAI Workshop on AI Planning for Cyber-Physical Systems (CAIPI)}},
 doi = {10.48550/arXiv.2312.08801},
 year = {2024}
}

@book{Ghallab.2025,
 author = {Ghallab, Malik and Nau, Dana S. and Traverso, Paolo},
 year = {2025},
 title = {{Acting, planning, and learning}},
 address = {Cambridge and New York, NY and Port Melbourne and New Delhi and Singapore},
 publisher = {{Cambridge University Press}},
 isbn = {9781009579346},
 doi = {10.1017/9781009579346}
}

@inproceedings{Wei.2022,
 author = {Wei, Jason and Wang, Xuezhi and Schuurmans, Dale and Bosma, Maarten and ichter, brian and Xia, Fei and Chi, Ed and Le, Quoc V. and Zhou, Denny},
 title = {{Chain-of-Thought Prompting Elicits Reasoning in Large Language Models}},
 volume = {35},
 publisher = {{Curran Associates, Inc}},
 editor = {Koyejo, S. and Mohamed, S. and Agarwal, A. and Belgrave, D. and Cho, K. and Oh, A.},
 booktitle = {{Advances in Neural Information Processing Systems}},
 year = {2022}
}

@inproceedings{Hao.2023,
 author = {Hao, Shibo and Gu, Yi and Ma, Haodi and Hong, Joshua and Wang, Zhen and Wang, Daisy and Hu, Zhiting},
 title = {{Reasoning with Language Model is Planning with World Model}},
 pages = {8154--8173},
 publisher = {{Association for Computational Linguistics}},
 editor = {Bouamor, Houda and Pino, Juan and Bali, Kalika},
 booktitle = {{Proceedings of the 2023 Conference on Empirical Methods in Natural Language Processing}},
 address = {Stroudsburg, PA, USA},
 doi = {10.18653/v1/2023.emnlp-main.507}, 
 year = {2023}
}

@inproceedings{Wei.2025,
 author = {Wei, Hui and Zhang, Zihao and He, Shenghua and Xia, Tian and Pan, Shijia and Liu, Fei},
 title = {{PlanGenLLMs: A Modern Survey of LLM Planning Capabilities}},
 pages = {19497--19521},
 publisher = {{Association for Computational Linguistics}},
 editor = {Che, Wanxiang and Nabende, Joyce and Shutova, Ekaterina and Pilehvar, Mohammad Taher},
 booktitle = {{Proceedings of the 63rd Annual Meeting of the Association for Computational Linguistics (Volume 1: Long Papers)}},
 address = {Vienna, Austria},
 doi = {10.18653/v1/2025.acl-long.958}, 
 year = {2025}
}

@article{Pallagani.2024,
 author = {Pallagani, Vishal and Muppasani, Bharath Chandra and Roy, Kaushik and Fabiano, Francesco and Loreggia, Andrea and Murugesan, Keerthiram and Srivastava, Biplav and Rossi, Francesca and Horesh, Lior and Sheth, Amit},
 year = {2024},
 title = {{On the Prospects of Incorporating Large Language Models (LLMs) in Automated Planning and Scheduling (APS)}},
 pages = {432--444},
 volume = {34},
 issn = {2334-0843},
 journal = {{Proceedings of the International Conference on Automated Planning and Scheduling}},
 doi = {10.1609/icaps.v34i1.31503}
}

@article{Fikes.1971,
 author = {Fikes, Richard E. and Nilsson, Nils J.},
 year = {1971},
 title = {{Strips: A new approach to the application of theorem proving to problem solving}},
 pages = {189--208},
 volume = {2},
 number = {3-4},
 issn = {00043702},
 journal = {{Artificial Intelligence}},
 doi = {10.1016/0004-3702(71)90010-5}
}

@article{Ghallab.1998,
 author = {Ghallab, Malik and Knoblock, Craig and Wilkins, David and Barrett, Anthony and Christianson, Dave and Friedman, Marc and Kwok, Chung and Golden, Keith and Penberthy, Scott and Smith, David and Sun, Ying and Weld, Daniel},
 year = {1998},
 title = {{PDDL - The Planning Domain Definition Language}},
 journal = {{Golden, S. Penberthy, DE Smith, Y. Sun, D. Weld, and D. Mcdermott, The planning domain definition language, Yale CVC, Tech. Rep}}
}

@article{Cashmore.2020,
 author = {Cashmore, Michael and Magazzeni, Daniele and Zehtabi, Parisa},
 year = {2020},
 title = {{Planning for Hybrid Systems via Satisfiability Modulo Theories}},
 pages = {235--283},
 volume = {67},
 journal = {{Journal of Artificial Intelligence Research}},
 doi = {10.1613/jair.1.11751}
}

@article{KBH+_AReferenceModelfor_15.09.2022b,
 author = {K{\"o}cher, Aljosha and Belyaev, Alexander and Hermann, Jesko and Bock, J{\"u}rgen and Meixner, Kristof and Volkmann, Magnus and Winter, Michael and Zimmermann, Patrick and Grimm, Stephan and Diedrich, Christian},
 year = {2023},
 title = {{A reference model for common understanding of capabilities and skills in manufacturing}},
 pages = {94--104},
 volume = {71},
 issn = {0178-2312},
 journal = {{at - Automatisierungstechnik}},
 shorthand = {KBH+23},
 doi = {10.1515/auto-2022-0117},
 number = {2},
}

@inproceedings{Wang.2023,
 author = {Wang, Zihao and Cai, Shaofei and Chen, Guanzhou and Liu, Anji and Ma, Xiaojian and Liang, Yitao and CraftJarvis, Team},
 title = {{Describe, explain, plan and select: interactive planning with large language models enables open-world multi-task agents}},
 publisher = {{Curran Associates Inc}},
 series = {{NIPS '23}},
 booktitle = {{Proceedings of the 37th International Conference on Neural Information Processing Systems}},
 year = {2023},
 address = {Red Hook, NY, USA}
}

@inproceedings{Zhou.2024,
 author = {Zhou, Andy and Yan, Kai and Shlapentokh-Rothman, Michal and Wang, Haohan and Wang, Yu-Xiong},
 title = {{Language agent tree search unifies reasoning, acting, and planning in language models}},
 publisher = {JMLR.org},
 series = {{ICML'24}},
 booktitle = {{Proceedings of the 41st International Conference on Machine Learning}},
 year = {2024}
}

@article{Silver.2024,
 author = {Silver, Tom and Dan, Soham and Srinivas, Kavitha and Tenenbaum, Joshua B. and Kaelbling, Leslie and Katz, Michael},
 year = {2024},
 title = {{Generalized Planning in PDDL Domains with Pretrained Large Language Models}},
 pages = {20256--20264},
 volume = {38},
 number = {18},
 issn = {2159-5399},
 journal = {{Proceedings of the AAAI Conference on Artificial Intelligence}},
 doi = {10.1609/aaai.v38i18.30006}
}

@misc{Liu.2023,
  author = {Liu, Bo and Jiang, Yuqian and Zhang, Xiaohan and Liu, Qiang and Zhang, Shiqi and Biswas, Joydeep and Stone, Peter},
  title  = {{LLM+P: Empowering Large Language Models with Optimal Planning Proficiency}},
  year   = {2023},
  note   = {arXiv:2304.11477 [cs.AI]},
  url    = {https://arxiv.org/abs/2304.11477}
}

@inproceedings{Ye.2023,
 author = {Ye, Xi and Chen, Qiaochu and Dillig, Isil and Durrett, Greg},
 title = {{SatLM: Satisfiability-Aided Language Models Using Declarative Prompting}},
 volume = {36},
 publisher = {{Curran Associates, Inc}},
 editor = {{A. Oh} and {T. Naumann} and {A. Globerson} and {K. Saenko} and {M. Hardt} and {S. Levine}},
 booktitle = {{Advances in Neural Information Processing Systems}},
 year = {2023}
}

@inproceedings{Rogalla.2017,
 author = {Rogalla, Antje and Niggemann, Oliver},
 title = {{Automated process planning for cyber-physical production systems}},
 pages = {1--8},
 publisher = {IEEE},
 isbn = {978-1-5090-6505-9},
 booktitle = {{2017 22nd IEEE International Conference on Emerging Technologies and Factory Automation}},
 year = {2017},
 address = {Piscataway, NJ},
 doi = {10.1109/ETFA.2017.8247683}
}

@inproceedings{Pan.2023,
 author = {Pan, Liangming and Albalak, Alon and Wang, Xinyi and Wang, William},
 title = {{Logic-LM: Empowering Large Language Models with Symbolic Solvers for Faithful Logical Reasoning}},
 pages = {3806--3824},
 publisher = {{Association for Computational Linguistics}},
 editor = {Bouamor, Houda and Pino, Juan and Bali, Kalika},
 booktitle = {{Findings of the Association for Computational Linguistics: EMNLP 2023}},
 address = {Stroudsburg, PA, USA},
 doi = {10.18653/v1/2023.findings-emnlp.248},
 year = {2023}
}

@inproceedings{Jha.2023,
 author = {Jha, Sumit Kumar and Jha, Susmit and Lincoln, Patrick and Bastian, Nathaniel D. and Velasquez, Alvaro and Ewetz, Rickard and Neema, Sandeep},
 title = {{Counterexample Guided Inductive Synthesis Using Large Language Models and Satisfiability Solving}},
 pages = {944--949},
 publisher = {IEEE},
 isbn = {979-8-3503-2181-4},
 booktitle = {{MILCOM 2023 - 2023 IEEE Military Communications Conference (MILCOM)}},
 year = {2023},
 address = {Piscataway, NJ},
 doi = {10.1109/MILCOM58377.2023.10356332}
}

@misc{Hirsch.2024,
 author = {Hirsch, Eran and Uziel, Guy and Anaby-Tavor, Ateret},
 year = {2024},
 title = {{What's the Plan? Evaluating and Developing Planning-Aware Techniques for Language Models}},
 note   = {arXiv:2402.11489 [cs.CL]},
 url = {https://arxiv.org/abs/2402.11489}
}

@article{Li.2024,
 author = {Li, Xinyi and Wang, Sai and Zeng, Siqi and Wu, Yu and Yang, Yi},
 year = {2024},
 title = {{A survey on LLM-based multi-agent systems: workflow, infrastructure, and challenges}},
 volume = {1},
 number = {1},
 journal = {{Vicinagearth}},
 doi = {10.1007/s44336-024-00009-2}
}

@inproceedings{Yao.2023,
 author = {Yao, Shunyu and Zhao, Jeffrey and Yu, Dian and Du, Nan and Shafran, Izhak and Narasimhan, Karthik and Cao, Yuan},
 title = {{ReAct: Synergizing Reasoning and Acting in Language Models}},
 booktitle = {{International Conference on Learning Representations}},
 year = {2023}
}

\end{document}